%% file: example_paper.tex
\documentclass{article}

\usepackage{microtype}
\usepackage{graphicx}
\usepackage{subfigure}
\usepackage{booktabs} 
\usepackage{array}
\usepackage{multirow}
\usepackage{makecell}
\usepackage{enumitem}
\usepackage{hyperref}

\usepackage[accepted]{icml2024}

\usepackage{amsmath}
\usepackage{amssymb}
\usepackage{mathtools}
\usepackage{amsthm}

\usepackage[capitalize,noabbrev]{cleveref}

\theoremstyle{plain}

\theoremstyle{definition}

\theoremstyle{remark}

\usepackage[textsize=tiny]{todonotes}

\definecolor{darkpastelgreen}{rgb}{0.01, 0.75, 0.24}

\icmltitlerunning{Faithful and Fast Influence Function via Advanced Sampling}

\begin{document}

\twocolumn[
\icmltitle{Faithful and Fast Influence Function via Advanced Sampling}



\icmlsetsymbol{equal}{*}

\begin{icmlauthorlist}
\icmlauthor{Jungyeon Koh}{1}
\icmlauthor{Hyeonsu Lyu}{1}
\icmlauthor{Jonggyu Jang}{2}
\icmlauthor{Hyun Jong Yang}{3}
\end{icmlauthorlist}

\icmlaffiliation{1}{Department of Electrical Engineering, Pohang University of Science and Technology, Pohang, Republic of Korea}
\icmlaffiliation{2}{Department of Electrons Engineering, Chungnam University, Daejeon, Republic of Korea}
\icmlaffiliation{3}{Department of Electrical and Computer Engineering, Seoul National University, Seoul, Republic of Korea}

\icmlcorrespondingauthor{Hyun Jong Yang}{hjyang@snu.ac.kr}

\icmlkeywords{Machine Learning, ICML}

\vskip 0.3in
]



\printAffiliationsAndNotice{}

\begin{abstract}
\textit{How can we explain the influence of training data on black-box models?} 
Influence functions (IFs) offer a post-hoc solution by utilizing gradients and Hessians. 
However, computing the Hessian for an entire dataset is resource-intensive, necessitating a feasible alternative.
A common approach involves randomly sampling a small subset of the training data, but this method often results in  highly inconsistent IF estimates due to the high variance in sample configurations. 
To address this, we propose two advanced sampling techniques based on features and logits. 
These samplers select a small yet representative subset of the entire dataset by considering the stochastic distribution of features or logits, thereby enhancing the accuracy of IF estimations. 
We validate our approach through class removal experiments, a typical application of IFs, using the \(\mathrm{F}_1\)-score to measure how effectively the model forgets the removed class while maintaining inference consistency on the remaining classes. 
Our method reduces computation time by 30.1\% and memory usage by 42.2\%, or improves the \(\mathrm{F}_1\)-score by 2.5\% compared to the baseline. 

\end{abstract}

\section{Introduction}

A comprehensive understanding of model behaviors has become paramount, particularly as ensuring alignment with human ethics and societal values emerges as a critical concern in the renaissance of hyper-scale models.
The recent technical report \cite{park2024-AiDeception} exemplified such concerns, addressing the potential of AI to deceive humans.
However, the paradigm shift towards deeper and larger architectures has posed significant challenges for providing explainability and interpretability.

Influence functions---originating from classical statistics \cite{hampel1974influence}---have revived as a crucial breakthrough in enhancing the transparency and accessibility of black-box AI models \cite{koh2017understanding}.
For black-box AI models, influence functions provide a direct evaluation of how the inclusion or exclusion of data affects model parameters by leveraging only their Hessians and gradients.
Thus, unlike traditional retraining-based analyses, such as leave-$k$-out validations, influence functions significantly reduce the costs associated with fine-tuning and retraining, thereby mitigating the carbon footprint of model analysis~\citep{koh2019accuracy}.
Recent studies have demonstrated the robustness of influence functions across various domains, including model analysis \cite{koh2017understanding, kong2022-Relabeling}, priori and post-hoc data processing \cite{Lee2020-DataAug_IF, yang2023-dataset, cohen2020detecting}, machine unlearning \cite{grosse2023studying}, and natural language processing \citep{jain2022-SeqeunceTagging, ye2022-progen}.

While influence functions have advanced the strive for explainability in black-box model inference, they encounter two major limitations: (1) high memory and computational demands, and (2) imprecise approximations when applied to large-scale models due to the theoretical necessity for Hessian inversion. The emergence of hyperscale AI systems, such as large language models (LLMs), has worsened these issues, posing further challenges to achieving practical real-world applications. To avoid such inefficient computations, primitive influence function methods employ random sampling when computing Hessians, which still fails to accurately estimate the true leave-one-out (LOO) effect.  

\begin{figure*}[h]
  \centering
  \includegraphics[width=0.82\linewidth]{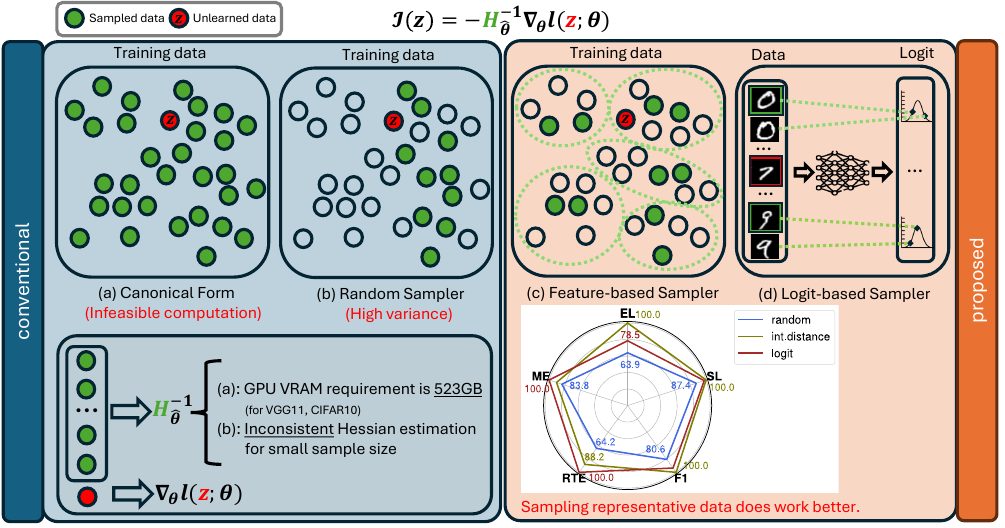}
  \caption{Overview of our approach. A quick evaluation shows the performance of three sampling methods under five metrics: exclusive-loss (EL), self-loss (SL), F1 score (F1), run-time efficiency (RTE), and memory efficiency (ME). Results show that improved samplings lead to more accurate estimations of unlearning effects within less memory and time. The influence functions require Hessian matrix of the \textcolor{darkpastelgreen}{sampled training dataset} and gradient vector of the \textcolor{red}{target data}. In \textcolor{blue}{conventional methods}, the Hessian matrix is (a) intractable or (b) possibly unreliable. In our method, \textcolor{orange}{advanced samplers} can choose a small but representative subset based on (c) feature and (d) logits. 
  }
  \label{fig:1}
\end{figure*}

To resolve these challenges, we propose advanced sampling techniques designed to preserve the accuracy of influence functions while enhancing computational and memory efficiency. Figure \ref{fig:1} illustrates the differences between our sampling methods and conventional random sampling. Our findings confirm that employing representative data points in Hessian computations improves both the accuracy and efficiency of influence function estimations.

\section{Revisiting Influence Functions}

\paragraph{Definitions and Implications.} Given a model \(\theta\) of size \(p\) in the parameter space \(\Theta\in\mathbb{R}^p\), and \(n\) training points \(z_1, .., z_n \in Z\), the empirical risk is defined as \(L(\theta) = \frac{1}{n} \sum_{i=1}^n l(z_i;\theta)\). Accordingly, the empirical risk minimizer is \(\hat{\theta} = \arg\min_{\theta \in \Theta} L(\theta)\).

Now, influence functions compute the parameter change if a certain data point \(z\) is upweighted by some small \(\epsilon\). Hence, an \(\epsilon\)-upweighted empirical risk minimizer is defined as
\begin{equation}
    \hat{\theta}_{\epsilon, z} = \arg\min_{\theta \in \Theta} L(\theta) + \epsilon l(z;\theta).
\end{equation}
Influence functions can be derived by using a Taylor expansion and a single Newton step as follows:
\begin{equation}\label{vanillaIF}
    \mathcal{I}(z):=\frac{d(\hat{\theta}_{\epsilon,z} - \hat{\theta})}{d\epsilon} \Bigg|_{\epsilon=0} = -\mathbf{H}_{\hat{\theta}}^{-1} \nabla_\theta l(z;\theta),
\end{equation}
where the Hessian is \(\mathbf{H}_{\hat{\theta}}:=\frac{1}{n} \sum_{i=1}^n \nabla_\theta^2 l(z_i;\hat{\theta}) \in \mathbb{R}^{p\times p}\).

IFs align well with LOO retraining for linear models, but \citet{koh2019accuracy, basu2020influence} revealed that this breaks down when applied to larger datasets or deeper models. This discrepancy arises from a strong convexity assumption, which is often violated in modern deep neural networks. 
Moreover, \citet{bae2022if} showed that IFs align better with the proximal Bregman response function (PBRF), which approximates the effect of removing a data point while preserving prediction consistency on the remaining dataset.
Since PBRF can effectively address questions about model behaviors, IFs remain a valuable post-hoc analysis tool, serving as a good approximation of PBRF.

\paragraph{LiSSA for a faster computation.}
Given \(p=\vert\Theta\vert\), inverting $p\times p$ Hessian as in \eqref{vanillaIF} imposes a huge computational bottleneck with a complexity of \(O(p^3)\). 
Accordingly, \citet{koh2017understanding} employed an iterative method using LiSSA \cite{agarwal2017second} to compute the inverse-Hessian-vector product (iHVP) instead of directly inverting the Hessians. The iterative approximation can be represented as 
\begin{equation}
    \mathcal{I}_k = \mathcal{I}_0 + (\mathbf{I} - \mathbf{H_{\hat{\theta}}})\mathcal{I}_{k-1},
\end{equation}
where index \(k\) indicates the timestep in this iterative process and \(\mathcal{I}_0 = \nabla l(z;\hat{\theta})\).
\(\mathbf{H_{\hat{\theta}}}\) is estimated using \(\nabla^2 l(z_{s_i};\hat{\theta})\) from \(t\) randomly selected data samples \(z_{s_1},...,z_{s_t}\). 
This recursive series converges to \(\mathcal{I}(z)\) as \(k\to \infty\) based on the validity of the Taylor expansion.
The iteration stops when \(\Vert \mathcal{I}_{k+1} - \mathcal{I}_{k}\Vert \leq \delta\) for a predefined threshold \(\delta\).

\paragraph{Shortcomings of LiSSA.}
The LiSSA iteration tends to produce inaccurate influence estimations. Additionally, it has a time complexity of \(O(nrp)\) for $n$ data points, $r$ iterations, and $p$ parameters.
To address these inherent challenges, \citet{basu2020second, yeh2022first, koh2017understanding} have suggested ``optimizing'' the computational budget associated with $n$ and $p$ by sampling the dataset and freezing network layers, which still fail to enhance accuracy. This finding aligns with \citet{feldman2020neural}, who confirmed that estimation errors can occur even in simple single-layer networks. 
Moreover, \citet{basu2020second} employed a second-order approximation, and \citet{teso2021interactive} used a Fisher information matrix \cite{lehmann2006theory} to improve accuracy. Nevertheless, both approaches endure a sharp increase in computational complexity, making them impractical for real-world applications.

Conversely, we believe that random sampling is responsible for inaccurate and unreliable LiSSA iterations. This is due to the high variance of the average loss associated with a limited number of sampling procedures and sampled instances. While expected Hessians and gradients from randomly sampled points are theoretically unbiased, the practical implementations suffers from this variance.

\section{Sampling Methods}
We assume that sampling representative data could enhance the accuracy and consistency of computing \(\mathcal{I}_k\), thereby reducing the required iterations. 
Influence functions mostly rely on random sampling to estimate $\mathbf{H}_{\hat{\theta}}\approx \mu(\nabla^2 l(z_{s_i};\hat{\theta}))$ \cite{koh2017understanding}, which suffers from high variance.
Conversely, employing advanced sampling methods could yield a more robust Hessian approximation. 
In essence, we aim to ``optimize'' the computational complexity in terms of $r$ by expediting the convergence of the LiSSA algorithm. 

In this section, we introduce several novel sampling methods based on the features and logits of the training data. 

\paragraph{Feature-based sampling.}
We assume that organizing data points within a latent feature space and selecting samples based on the space topology can avoid the unexpected variance of random samplers. 
Hence, we extract features in an extrinsic and intrinsic manner, then sample the features using two sampling methods.

We adopt a pre-trained Vision Transformer (ViT) model \cite{dosovitskiy2020image} as an extrinsic feature extractor because the ViT is well-known for its effectiveness in extracting general features in diverse network architectures. However, the pre-trained ViT model takes additional time for fine-tuning; and we cannot tell how much the sampling contributes to the accuracy of influence functions when employing additional model in estimating influence functions.
Accordingly, as part of an ablation approach, we design an intrinsic feature extractor, which directly uses the network being investigated to avoid transfer of trust problems.

Thereafter, we develop two sampling methods using both extrinsic and intrinsic feature extractors as follows:
\begin{itemize}
    \vspace{-0.1cm}
    \item \textbf{Top-\(k\) sampling}: Compute \(C\) centroids in the feature space using the K-means algorithm for a pre-defined \(C\). Then, select \(k\) samples that are the nearest to each centroid, resulting in a total selection of \(kC\) samples.

    \item \textbf{Distance-weighted sampling} \cite{wu2017sampling}: For each extracted feature \(z_i\) and centroid \(c\), compute the \(l_2\) distance \(d_{z_i,c}\) and create a multinomial distribution with probability 
    \begin{equation}
        p_{z_i,c} = \frac{1}{d_{z_i,c} - (\min_{z\in\mathbf{z}} d_{z,c} - \epsilon)},
    \end{equation}
    for \(\epsilon > 0\). A larger \(\epsilon\) increases the probability for data points farther from the centroid,  adding a certain degree of stochasticity compared to top-\(k\) sampling. Then, select \(k\) samples from the multinomial distribution for each centroid \(c\), again resulting in \(kC\) samples in total. 
\end{itemize}

Combining the two feature extractors with the two sampling methods described above, we have four feature-based samplers as follows: extrinsic Top-\(k\) sampling (\textbf{ext. top-k}), intrinsic Top-\(k\) sampling (\textbf{int. top-k}), extrinsic distance-weighted sampling (\textbf{ext. distance}), and intrinsic distance-weighted sampling (\textbf{int. distance}).

\paragraph{Logit-based sampling.} We design another logit-based sampler (\textbf{logit}) based on a class-wise softmax score of each data point \(x_i\) across \(Y\) classes. This involves creating a multinomial distribution for each class \(y\in \mathbf{y}\) with the probability of 
\begin{equation}
   p_{x_i, y}= [\text{softmax}(x_i;\theta)]_y,
\end{equation}
Then, \(k\) samples are chosen for each class \(y\) from the multinomial, resulting in \(kY\) samples in total.

\section{Experiments}
We evaluate the efficacy of our sampling methods by performing a class removal task using the original influence function \citep{koh2017understanding}. 



The experiments are performed on VGG11 \cite{simonyan2014very} trained with CIFAR-10 as described in \cite{koh2017understanding, lyu2024deeper}, and training points labeled as ``8'' (horse) are removed. In addition, we evaluate the sampling methods on class removal tasks with the other influence alternatives \cite{agarwal2017second, guo2020fastif, schioppa2022scaling, lyu2024deeper} and datasets in Appendix~\ref{appendix:A}.

\begin{figure*}[h]
  \includegraphics[width=0.49\linewidth]{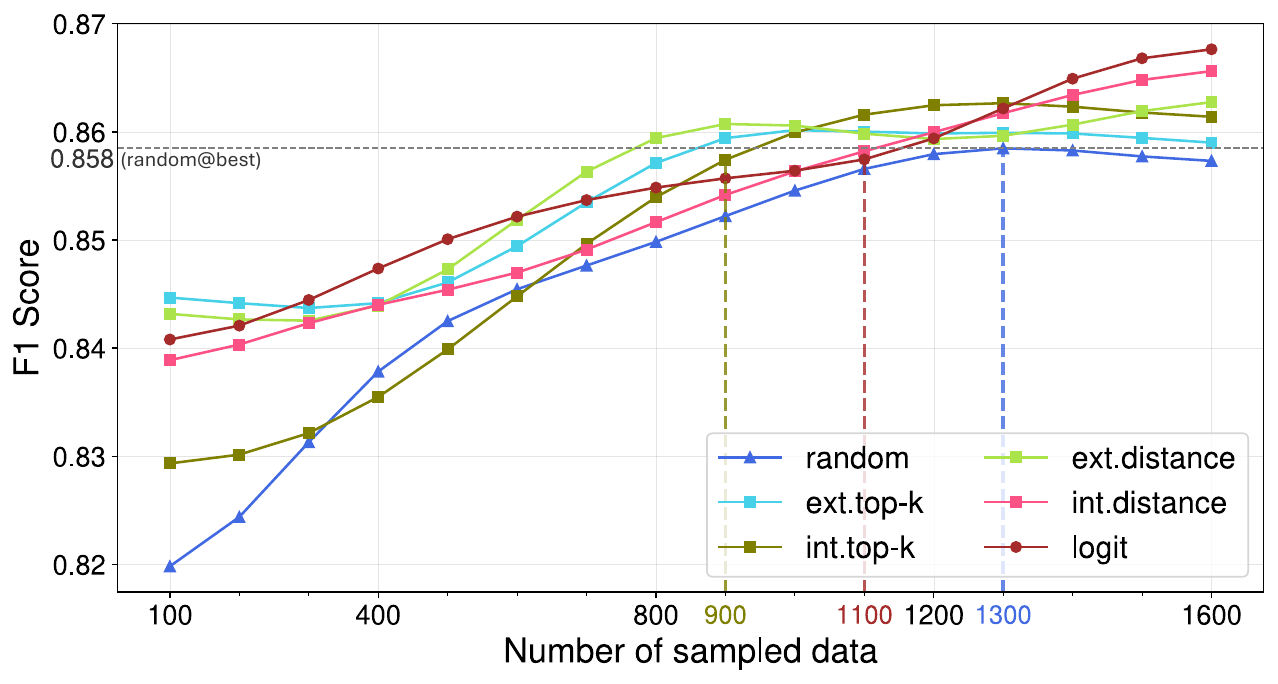}
  \includegraphics[width=0.49\linewidth]{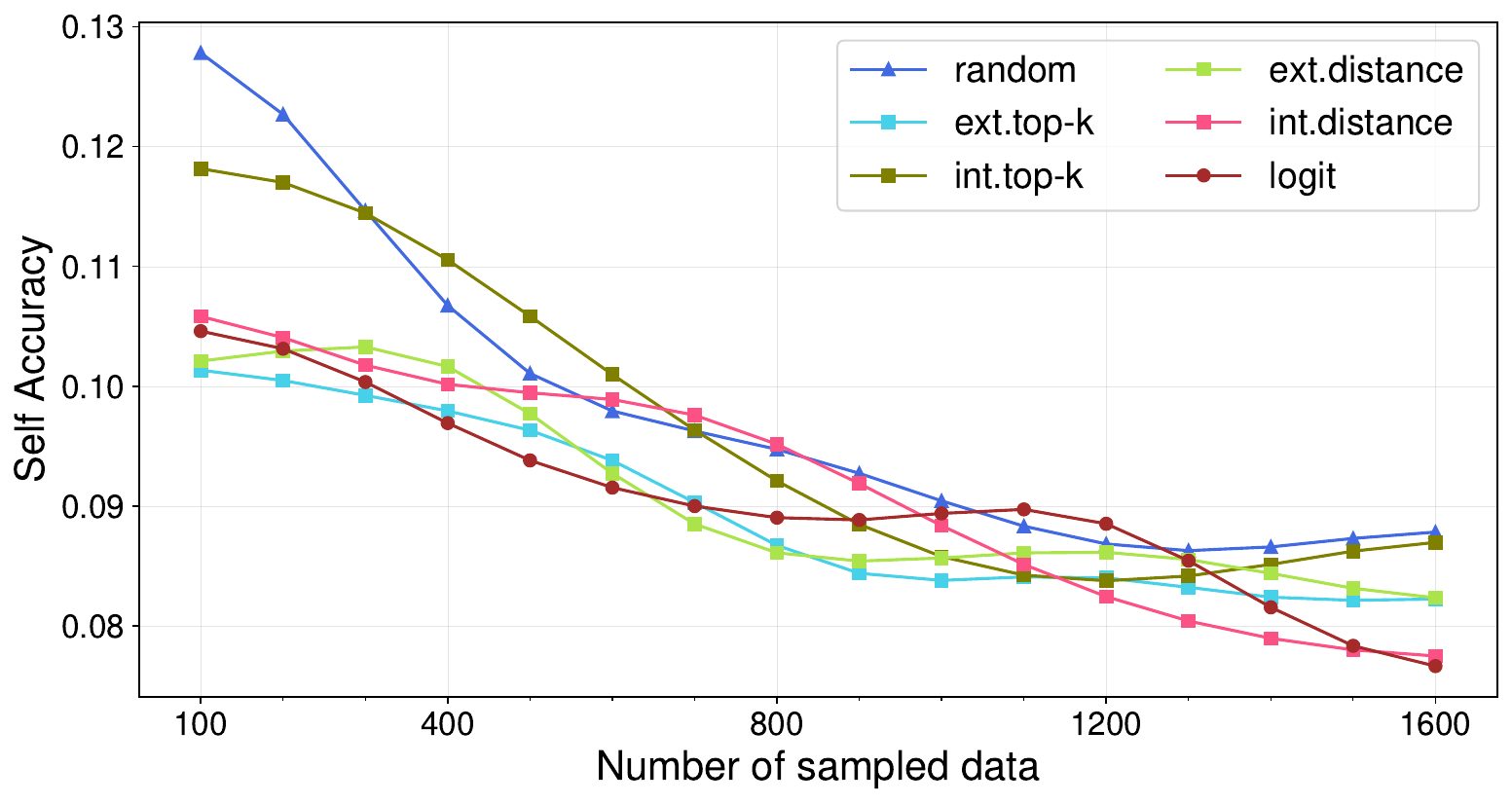}
  \includegraphics[width=0.49\linewidth]{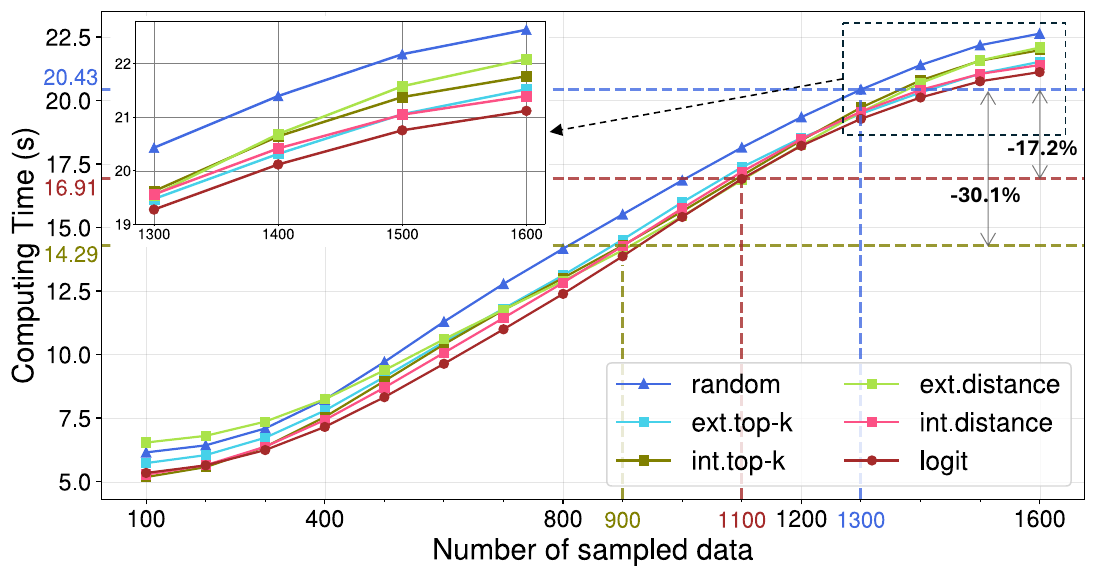}
  \includegraphics[width=0.49\linewidth]{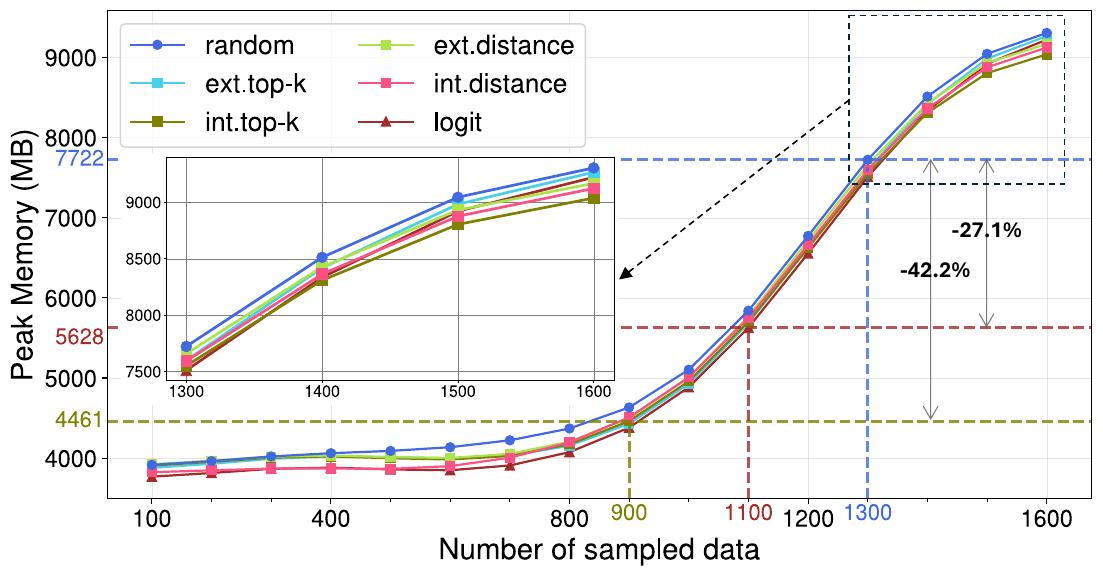}
  \vspace{-0.25cm}
  \caption{Evaluation results on the class removal task for VGG11 with CIFAR-10. Lower values indicate better performance, except for \(\mathrm{F}_1\)-score.}
  \label{fig:2}
\end{figure*}
\vspace{-.1cm}

\paragraph{Evaluation metrics.} We outline the following metrics to evaluate the accuracy and computational efficiency of the sampling methods.
\begin{itemize}
    \item Self-loss (\(\mathrm{SL}\)): The loss for the removed data, denoted as \(\sum_{z\in Z'}\ l(z;\theta)\), where \(Z'\) is a set of all removed data points.
    \item \(\mathrm{F}_1\)-score (\(\mathrm{F}_1\)): A modified \(\mathrm{F}_1\) score, incorporating self-accuracy (\(\mathrm{SA}\)) and exclusive-accuracy (\(\mathrm{EA}\)) as \(\mathrm{F}_1 = 2 \frac{\mathrm{EA}(1-\mathrm{SA})}{1+\mathrm{EA}-\mathrm{SA}}\).
    \item Run-time efficiency (\(\mathrm{RTE}\)): The average computing time until the influence function converges.
    \item Memory efficiency (\(\mathrm{ME}\)): The peak memory consumption while computing influence functions measured by monitoring memory usage.
\end{itemize}

\paragraph{Results.} 
Figure~\ref{fig:2} illustrates how the network behaves using the above metrics when the number of samples increases.
The graphs are obtained by averaging the results of 25 individual experiments.
Since the standard deviation (\(\mathrm{SD}\)) is another critical indicator to verify the faithfulness of the sampling methods, we provide the corresponding \(\mathrm{SD}\) of Fig.~\ref{fig:2} in Appendix \ref{appendix:C}.

We summarize the key findings as follows:
\begin{itemize}[leftmargin=.3cm]
\vspace{-0.1cm}
\item \textbf{The logit yields the most accurate estimation over other methods.} Notably, both the \(\mathrm{F}_1\)-score and self-accuracy significantly improve as the number of samples increases.
We believe the superior performance of the \textbf{logit} is likely due to its utilization of the entire neural network, unlike other sampling methods.
Remarkably, the \textbf{logit} takes the least compute cost as it just maps the softmax result without any intervention of external neural network or K-means algorithm.

\vspace{-0.1cm}
\item \textbf{Distance-weighted samplers perform slightly worse than the \textbf{logit}, but still show satisfactory results.}  The \textbf{int. distance} and \textbf{ext. distance} sampler also shows comparable results to the \textbf{logit} in both \(\mathrm{F}_1\)-score and self-accuracy. The result implies that distribution-based samples from \{\textbf{logit}, \textbf{int. distance}, \textbf{ext. distance}\} provide a more comprehensive representation of the entire dataset than the samples from deterministic samplers \{\textbf{int. distance}, \textbf{ext. distance}\} and \textbf{random}.

\vspace{-0.1cm}
\item \textbf{Intrinsic samplers outperform extrinsic samplers.} For both top-\(k\) and distance-weighted sampling, using the model itself as an intrinsic feature extractor yields more accurate estimations than employing an additional ViT model as an extrinsic feature extractor. It indicates that using the model being investigated for feature extraction more effectively represents the true feature space than relying on an external model.

\vspace{-0.1cm}
\item \textbf{The random gets comparable to the other samplers as the sample count increases.} 
This is natural as the sample data points become sufficiently representative even selected by the \textbf{random}.

\vspace{-0.1cm}
\item \textbf{Remarkably, the logit and int. top-$k$ greatly reduce both execution time and memory.} 
To achieve the best \(\mathrm{F}_1\)-score of the \textbf{random}, the number of samples required is 1,300 for the \textbf{random}.
Meanwhile, the \textbf{logit} and \textbf{int. top-$k$} only require 900 and 1,100 sample counts.
As a result, the \textbf{logit} and \textbf{int. top-$k$} save \(17.2\%\) and \(30.1\%\) in computing time, and \(22.3\%\) and \(40.6\%\) in memory, while maintaining the same performance.
\end{itemize}

\section{Discussion}

\paragraph{Summary.}
This paper deals with the challenge of efficient data sampling for computing influence functions on black-box AI models. 
Traditional methods relying on random sampling often produce inaccurate influence estimations due to high variance. 
To address this, we propose advanced samplers based on features and logits, selecting a representative subset of the dataset. 
Our experiments show that the proposed methods improve the accuracy of influence functions even with less time and memory usage.

\vspace{-0.3cm}
\paragraph{Limitation and future works.}
Our methods provide efficient sampling methods for estimating influence functions, which consistently outperform the baselines. 
However, our experimental analysis lacks a variety applications for influence functions. 
Also, as a future plan, we aim to explore the efficacy of sampling removal data in expediting class unlearning tasks. 
Furthermore, we plan to devise an effective update rule for influence functions with a much larger update rate than the theoretical value.

\vspace{-0.3cm}
\paragraph{Societal impact.}
AIs exhibit striking capabilities beyond our imagination, but their internal mechanisms remain obscure. Influence functions, which leverage training data to explain models, serve as a cornerstone of a bottom-up approach for mechanistic understanding of AI. We anticipate that influence functions contribute to building robust AI systems through a comprehensive understanding of the system. 

\section*{Acknowledgements}

This research was supported by the IITP(Institute for Information \& Communications Technology Planning \& Evaluation), grant funded by MSIT(Ministry of Science and ICT) (RS-2023-00229541, Development of Big Data and Artificial Intelligence Based Radio Monitoring Platform). This research was also supported by the MSIT(Ministry of Science and ICT), Korea, under the ITRC(Information Technology Research Center) support program(IITP-2024-2021-0-02048) supervised by the IITP(Institute for Information \& Communications Technology Planning \& Evaluation).

\section*{Impact Statement}

This paper aims to advance the field of machine learning. While there are many potential societal implications of our work, we do not believe any need to be specifically highlighted here.

\bibliography{example_paper}
\bibliographystyle{icml2024}


\newpage
\appendix
\onecolumn



\section{Accuracy Evaluation Details}
\label{appendix:A}

All experiments are performed on Linux with an NVIDIA Geforce RTX 3080Ti (12GB) GPU. 
CUDA version is 11.6 and Driver Version is 510.108.03. All codes are written under Python 3.10.10 and PyTorch 2.3.0.  

Altogether, for each dataset and model pair, both a target classification model and a ViT model, used as an extrinsic feature extractor, are initially well-trained. Subsequently, samplers generate sampled data based on its feature or logit distribution. To evaluate the performance of our sampling methods among different influence function approaches, we utilize five benchmarks: LiSSA-based IF (IF) \cite{koh2017understanding}, projected IF (PIF), generalized IF (GIF), freezed IF (FIF) \cite{lyu2024deeper}, and second-order IF (SIF) \cite{basu2020second}. Using these benchmarks alongside our sampling methods, we conduct a class removal experiment on: (1) Alexnet with the MNIST dataset, and (2) VGG11 with the CIFAR-10 dataset. The accuracy and consistency of our sampling methods on all datasets are presented in Table \ref{tab1} and Table \ref{tab2}, respectively. 

\input{table1.tex}
\input{table2.tex}

\section{Consistency Evaluation Details}
\label{appendix:C}

\begin{figure*}[h]
  \includegraphics[width=0.49\linewidth]{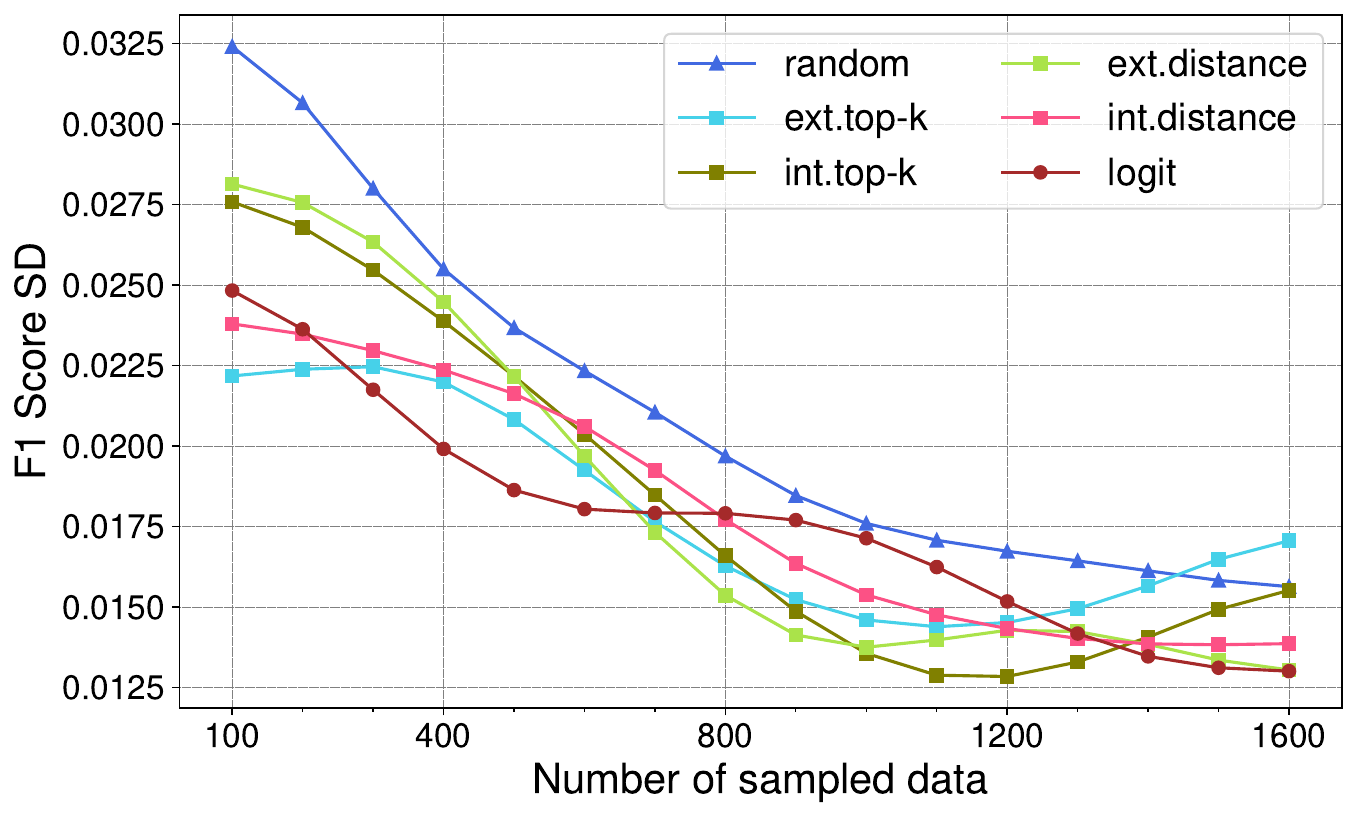}
  \includegraphics[width=0.49\linewidth]{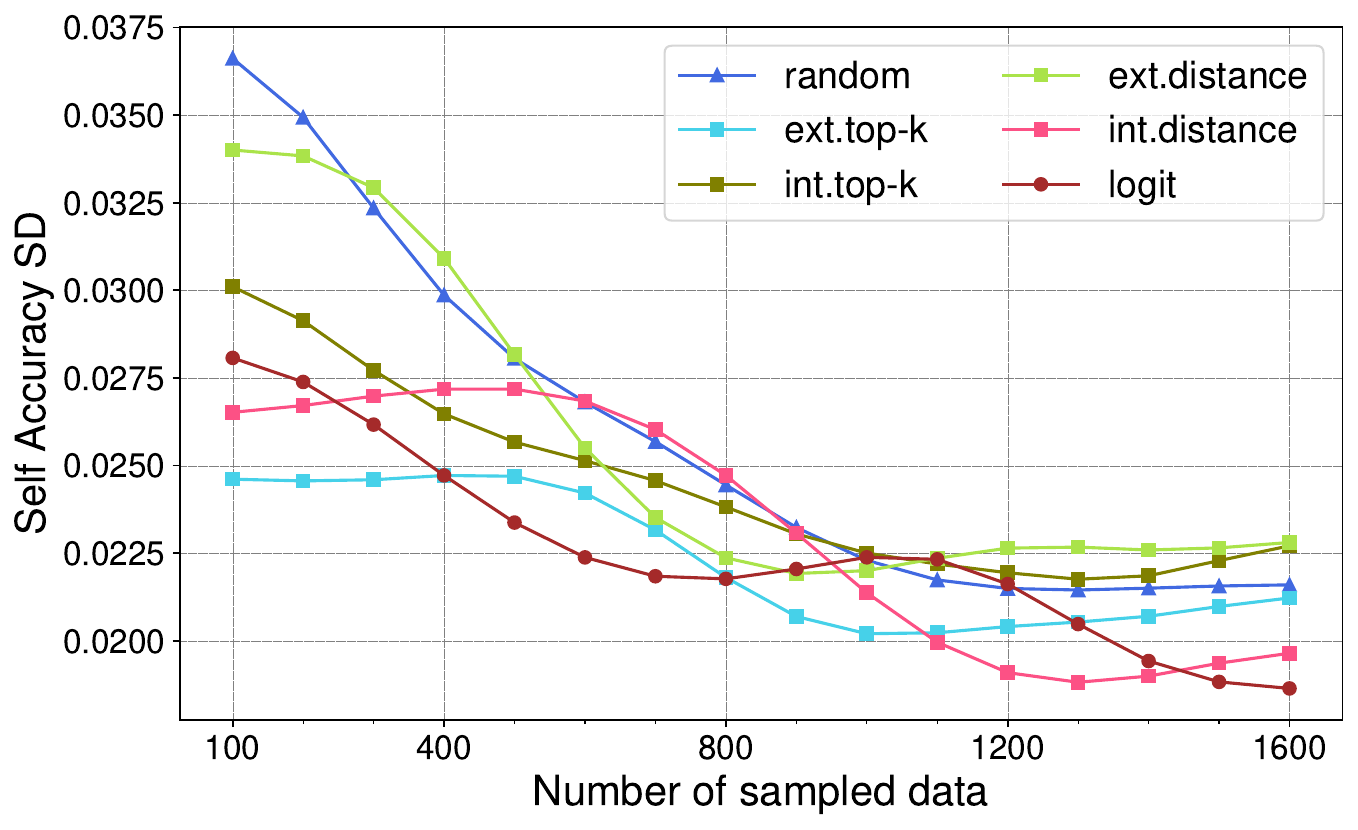}
  \includegraphics[width=0.49\linewidth]{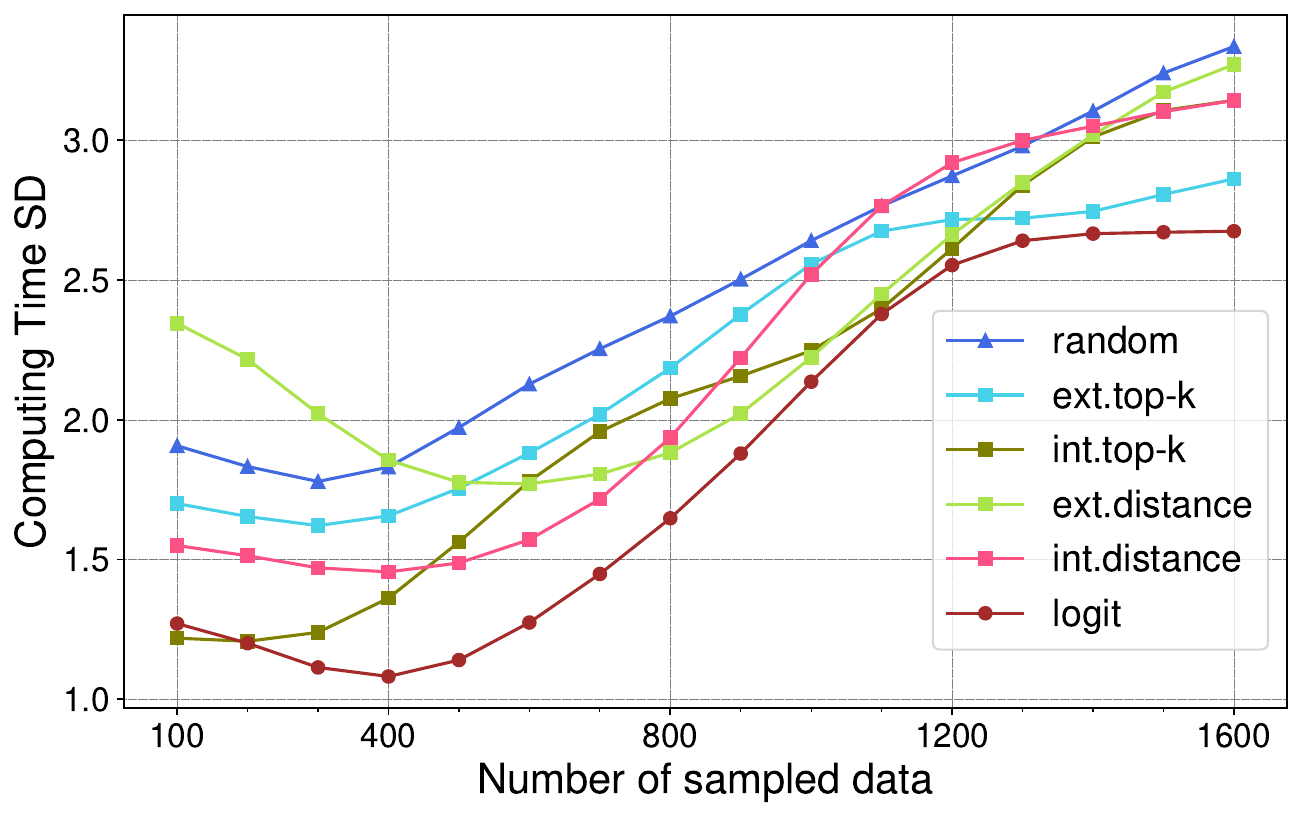}
  \includegraphics[width=0.49\linewidth]{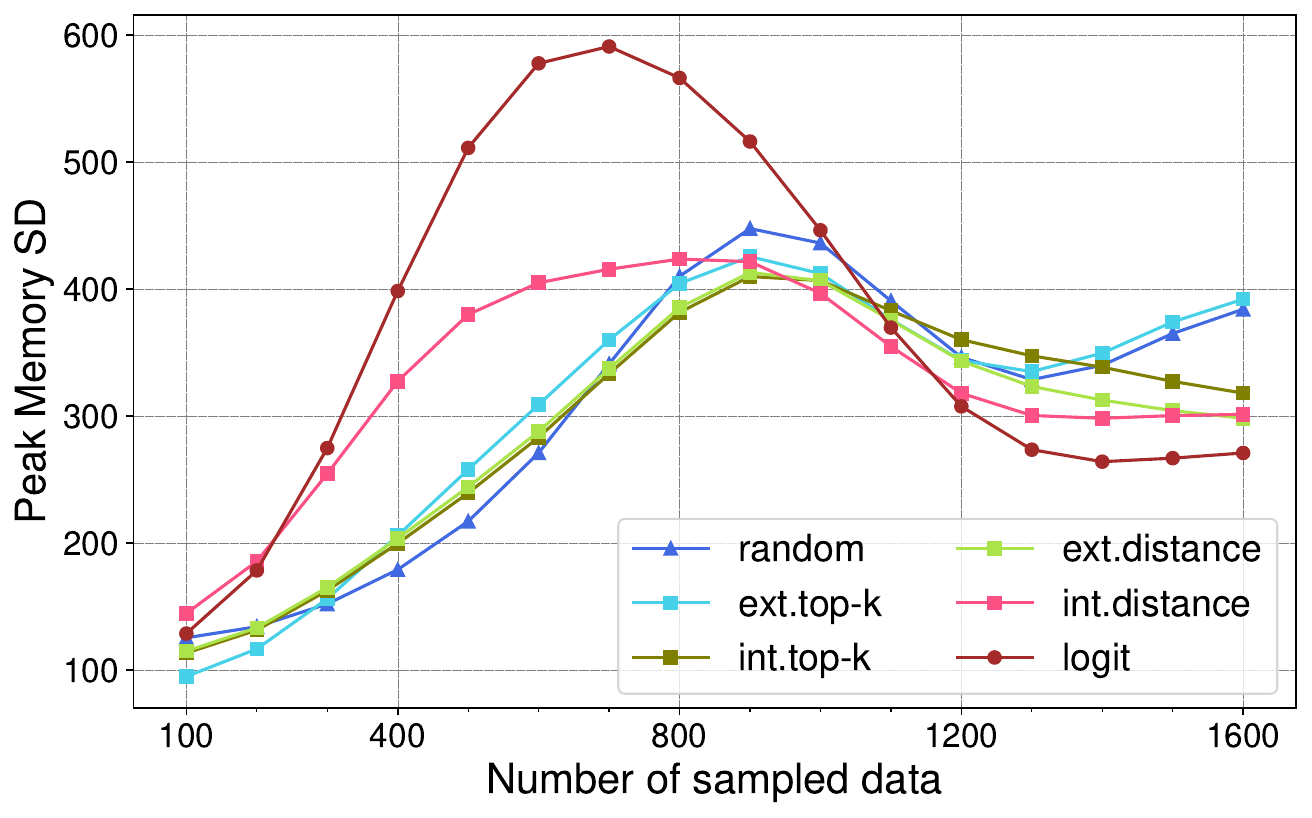}
  \vspace{-0.25cm}
  \caption{Standard deviation of evaluation metrics presented in Fig. \ref{fig:2}.}
  \label{fig:3}
\end{figure*}

We assess the standard deviation (\(\mathrm{SD}\)) of evaluation results to measure the consistency of influence function estimations with our novel sampling methods. 
The standard deviation 
of Fig. 2 is shown in Figure \ref{fig:3}. Based on this figure, the following observation can be additionally made:
\begin{itemize}
    \item \textbf{Proposed samplers provide more faithful evaluations with smaller deviations compared to the random sampler.} In particular, the logit-based sampler shows the smallest standard deviation, while extrinsic and intrinsic distance-weighted based samplers rank the second and third-smallest. This result strengthens our previous observation that stochastic sampling methods provide more accurate and consistent results than deterministic sampling methods.   
\end{itemize}

\end{document}

%% file: table1.tex
\begin{table}[h]
\centering
\caption{Overall performance benchmark of sampling methods for various influence function methods. The Alexnet model and the MNIST dataset are used. The best and the second-best performing methods are highlighted in bold, with the best-performing ones also marked with a superscript asterisk.}

\begin{tabular}{c c c c c c c c}
\Xhline{3\arrayrulewidth}
& & Random & Ext. Top-\(k\) & Int. Top-\(k\) & Ext. Distance & Int. Distance & Logit \\ \hline\hline
\multirow{3}{*}{IF} & EL($\downarrow$) 
& 0.051 \(\pm\) 0.008    
& 0.058 \(\pm\) 0.014    
& 0.063 \(\pm\) 0.021    
& 0.052 \(\pm\) 0.012    
& 0.055 \(\pm\) 0.008    
& 0.050 \(\pm\) 0.005 \\ 
                    & SL($\uparrow$) 
& 7.77 \(\pm\) 1.74 
& 8.71 \(\pm\) 2.62 
& 8.95 \(\pm\) 1.21 
& 7.72 \(\pm\) 1.82 
& 8.36 \(\pm\) 1.69 
& 7.93 \(\pm\) 0.85 \\ 
                    & F1($\uparrow$) 
& 0.963 \(\pm\) 0.016 
& 0.967 \(\pm\) 0.012 
& 0.967 \(\pm\) 0.012 
& 0.955 \(\pm\) 0.029 
& \textbf{0.968 \(\pm\) 0.011}
& \textbf{0.967 \(\pm\) 0.006}* \\ \hline

\multirow{3}{*}{PIF} & EL($\downarrow$) 
& 0.097 \(\pm\) 0.039 
& 0.076 \(\pm\) 0.043 
& 0.192 \(\pm\) 0.023 
& 0.066 \(\pm\) 0.014  
& 0.120 \(\pm\) 0.019 
& 0.078 \(\pm\) 0.047 \\
                    & SL($\uparrow$) 
& 8.95 \(\pm\) 2.30 
& 9.34 \(\pm\) 2.52 
& 8.21 \(\pm\) 1.12 
& 9.01 \(\pm\) 1.51 
& 9.73 \(\pm\) 1.14 
& 8.41 \(\pm\) 0.34 \\ 
                    & F1($\uparrow$) 
& 0.979 \(\pm\) 0.011 
& 0.964 \(\pm\) 0.046 
& 0.958 \(\pm\) 0.005 
& \textbf{0.990 \(\pm\) 0.002}*
& 0.981 \(\pm\) 0.003 
& \textbf{0.988 \(\pm\) 0.006} \\ \hline 

\multirow{3}{*}{FIF} & EL($\downarrow$) 
& 0.140 \(\pm\) 0.056 
& 0.085 \(\pm\) 0.036 
& 0.182 \(\pm\) 0.080 
& 0.066 \(\pm\) 0.027  
& 0.147 \(\pm\) 0.050 
& 0.082 \(\pm\) 0.040 \\
                    & SL($\uparrow$) 
& 9.88 \(\pm\) 1.98 
& 9.34 \(\pm\) 2.53 
& 8.91 \(\pm\) 1.47 
& 8.99 \(\pm\) 2.01 
& 9.89 \(\pm\) 0.97 
& 8.13 \(\pm\) 0.61 \\ 
                    & F1($\uparrow$) 
& 0.980 \(\pm\) 0.006 
& 0.960 \(\pm\) 0.050 
& 0.965 \(\pm\) 0.014 
& \textbf{0.989 \(\pm\) 0.004}*
& 0.979 \(\pm\) 0.003 
& \textbf{0.987 \(\pm\) 0.006} \\ \hline

\multirow{3}{*}{GIF} & EL($\downarrow$) 
& 0.100 \(\pm\) 0.031 
& 0.091 \(\pm\) 0.029 
& 0.152 \(\pm\) 0.065 
& 0.067 \(\pm\) 0.026  
& 0.126 \(\pm\) 0.026 
& 0.106 \(\pm\) 0.022 \\
                    & SL($\uparrow$) 
& 8.06 \(\pm\) 1.89 
& 9.16 \(\pm\) 2.18 
& 8.88 \(\pm\) 1.97 
& 9.55 \(\pm\) 1.84 
& 8.79 \(\pm\) 1.57 
& 8.12 \(\pm\) 0.85 \\ 
                    & F1($\uparrow$) 
& 0.978 \(\pm\) 0.010 
& 0.953 \(\pm\) 0.062 
& 0.956 \(\pm\) 0.019 
& \textbf{0.990 \(\pm\) 0.003}*
& 0.979 \(\pm\) 0.003 
& \textbf{0.982 \(\pm\) 0.002} \\ \hline

\multirow{3}{*}{SIF} & EL($\downarrow$) 
& 0.040 \(\pm\) 0.058 
& 0.084 \(\pm\) 0.044 
& 0.048 \(\pm\) 0.042 
& 0.076 \(\pm\) 0.008  
& 0.058 \(\pm\) 0.021 
& 0.020 \(\pm\) 0.024 \\
                    & SL($\uparrow$) 
& 10.05 \(\pm\) 2.39 
& 13.14 \(\pm\) 3.01 
& 10.44 \(\pm\) 2.81 
& 10.12 \(\pm\) 2.13
& 10.36 \(\pm\) 1.96 
& 10.50 \(\pm\) 1.68 \\ 
                    & F1($\uparrow$) 
& 0.948 \(\pm\) 0.055 
& \textbf{0.976 \(\pm\) 0.030}
& 0.958 \(\pm\) 0.047
& 0.951 \(\pm\) 0.034
& 0.958 \(\pm\) 0.081
& \textbf{0.978 \(\pm\) 0.030}* \\ 
\Xhline{3\arrayrulewidth}
\end{tabular}
\label{tab1}
\end{table}

%% file: table2.tex
\begin{table}[h]
\centering
\caption{Overall performance benchmark of sampling methods for various influence function methods. The VGG11 model and the CIFAR-10 dataset are used. The best and the second-best performing methods are highlighted in bold, with the best-performing ones also marked with a superscript asterisk.}

\begin{tabular}{c c c c c c c c}
\Xhline{3\arrayrulewidth}
& & Random & Ext. Top-\(k\) & Int. Top-\(k\) & Ext. Distance & Int. Distance & Logit \\ \hline\hline
\multirow{3}{*}{IF} & EL($\downarrow$) 
& 0.777 \(\pm\) 0.138    
& 0.646 \(\pm\) 0.089    
& 0.628 \(\pm\) 0.090    
& 0.614 \(\pm\) 0.078    
& 0.608 \(\pm\) 0.050    
& 0.678 \(\pm\) 0.070 \\ 
                    & SL($\uparrow$) 
& 7.83 \(\pm\) 1.27 
& 8.24 \(\pm\) 1.35 
& 8.04 \(\pm\) 1.15 
& 7.80 \(\pm\) 0.73 
& 8.14 \(\pm\) 0.32 
& 8.27 \(\pm\) 0.87 \\ 
                    & F1($\uparrow$) 
& 0.869 \(\pm\) 0.012 
& \textbf{0.874 \(\pm\) 0.018}
& 0.874 \(\pm\) 0.019
& 0.871 \(\pm\) 0.017
& 0.867 \(\pm\) 0.013
& \textbf{0.881 \(\pm\) 0.012} * \\ \hline

\multirow{3}{*}{PIF} & EL($\downarrow$) 
& 1.189 \(\pm\) 0.382 
& 1.580 \(\pm\) 0.194 
& 1.184 \(\pm\) 0.270 
& 1.245 \(\pm\) 0.162  
& 1.438 \(\pm\) 0.222 
& 0.936 \(\pm\) 0.081 \\
                    & SL($\uparrow$) 
& 7.93 \(\pm\) 1.10 
& 7.45 \(\pm\) 0.58 
& 8.56 \(\pm\) 0.97 
& 7.10 \(\pm\) 0.63 
& 8.24 \(\pm\) 1.03 
& 8.84 \(\pm\) 0.56 \\ 
                    & F1($\uparrow$) 
& 0.826 \(\pm\) 0.049 
& 0.781 \(\pm\) 0.030
& \textbf{0.828 \(\pm\) 0.047}
& 0.800 \(\pm\) 0.031
& 0.809 \(\pm\) 0.054 
& \textbf{0.863 \(\pm\) 0.015}* \\ \hline 

\multirow{3}{*}{FIF} & EL($\downarrow$) 
& 1.271 \(\pm\) 0.359 
& 0.980 \(\pm\) 0.227 
& 0.952 \(\pm\) 0.233 
& 0.916 \(\pm\) 0.170  
& 0.950 \(\pm\) 0.351 
& 1.154 \(\pm\) 0.185 \\
                    & SL($\uparrow$) 
& 7.04 \(\pm\) 1.44 
& 7.90 \(\pm\) 0.54 
& 7.72 \(\pm\) 1.14 
& 8.43 \(\pm\) 0.43 
& 7.98 \(\pm\) 0.93 
& 8.36 \(\pm\) 0.83 \\ 
                    & F1($\uparrow$) 
& 0.800 \(\pm\) 0.055 
& 0.852 \(\pm\) 0.025
& 0.857 \(\pm\) 0.038 
& \textbf{0.859 \(\pm\) 0.020}
& \textbf{0.860 \(\pm\) 0.043}*
& 0.835 \(\pm\) 0.029 \\ \hline

\multirow{3}{*}{GIF} & EL($\downarrow$) 
& 1.231 \(\pm\) 0.483 
& 1.261 \(\pm\) 0.261 
& 1.064 \(\pm\) 0.240 
& 1.175 \(\pm\) 0.256  
& 1.091 \(\pm\) 0.356 
& 1.030 \(\pm\) 0.344 \\
                    & SL($\uparrow$) 
& 7.33 \(\pm\) 1.14 
& 8.13 \(\pm\) 0.93 
& 8.28 \(\pm\) 0.68 
& 8.47 \(\pm\) 1.11 
& 7.87 \(\pm\) 0.92 
& 8.40 \(\pm\) 0.64 \\ 
                    & F1($\uparrow$) 
& 0.811 \(\pm\) 0.066 
& 0.814 \(\pm\) 0.039 
& \textbf{0.847 \(\pm\) 0.035} 
& 0.829 \(\pm\) 0.026
& 0.828 \(\pm\) 0.059 
& \textbf{0.852 \(\pm\) 0.048}* \\ \hline

\multirow{3}{*}{SIF} & EL($\downarrow$) 
& 0.984 \(\pm\) 0.242
& 1.297 \(\pm\) 0.228 
& 1.045 \(\pm\) 0.249
& 1.380 \(\pm\) 0.287  
& 0.987 \(\pm\) 0.341 
& 0.589 \(\pm\) 0.320 \\
                    & SL($\uparrow$) 
& 7.26 \(\pm\) 1.39 
& 6.95 \(\pm\) 1.26 
& 9.62 \(\pm\) 1.44 
& 6.90 \(\pm\) 1.28  
& 7.64 \(\pm\) 1.52 
& 8.11 \(\pm\) 1.30 \\ 
                    & F1($\uparrow$) 
& 0.839 \(\pm\) 0.052
& 0.819 \(\pm\) 0.060 
& \textbf{0.845 \(\pm\) 0.027}
& 0.834 \(\pm\) 0.059
& 0.826 \(\pm\) 0.032
& \textbf{0.897 \(\pm\) 0.021}* \\ \Xhline{3\arrayrulewidth}
\end{tabular}
\label{tab2}
\end{table}